\documentclass[letterpaper, 10 pt, conference]{ieeeconf}  

\IEEEoverridecommandlockouts                              

\usepackage{times}

\usepackage{multicol}
\usepackage[dvipsnames]{xcolor}
\usepackage{graphicx}
\usepackage{hyperref}
\usepackage{amsmath}
\usepackage{cleveref}
\usepackage[ruled,vlined,linesnumbered]{algorithm2e}
\usepackage[font=footnotesize,labelfont=bf]{caption}
\usepackage{subcaption}
\usepackage{bbm}
\usepackage{float}
\usepackage{color, soul} 
\usepackage{wrapfig,lipsum,booktabs}

\usepackage{array}
\usepackage[utf8]{inputenc}
\usepackage{amsmath, amssymb}
\usepackage{booktabs}
\usepackage{dsfont}
\usepackage{diagbox}
\usepackage{multirow}

\DeclareMathOperator*{\argmin}{arg\,min}

\usepackage{amsmath,amsfonts,bm}

\def\eqref#1{equation~\ref{#1}}

\def\1{\bm{1}}

\def\rmM{{\mathbf{M}}}

\DeclareMathAlphabet{\mathsfit}{\encodingdefault}{\sfdefault}{m}{sl}
\SetMathAlphabet{\mathsfit}{bold}{\encodingdefault}{\sfdefault}{bx}{n}

\def\gD{{\mathcal{D}}}

\def\gL{{\mathcal{L}}}

\def\gN{{\mathcal{N}}}

\newcommand{\E}{\mathbb{E}}

\newcommand{\btau}{\boldsymbol{\tau}}

\newcommand{\bJ}{\boldsymbol{J}}
\newcommand{\bO}{\boldsymbol{O}}
\newcommand{\bR}{\boldsymbol{R}}
\newcommand{\bI}{\boldsymbol{I}}

\newcommand{\bx}{\mathbf{x}}
\newcommand{\bo}{\mathbf{o}}
\newcommand{\br}{\mathbf{r}}
\newcommand{\bu}{\mathbf{u}}

\title{\LARGE \bf
Subgoal Diffuser: Coarse-to-fine Subgoal Generation to Guide Model Predictive Control for Robot Manipulation
}

\author{Zixuan Huang$^{1}$, Yating Lin$^{1}$, Fan Yang$^{1}$, Dmitry Berenson$^{1}$
\thanks{This work was supported in part by the Office of Naval Research Grant N00014-21-1-2118 and NSF grants IIS-1750489, IIS-2113401, and IIS-2220876. $^{1}$ Department of Robotics, University of Michigan, Ann Arbor}
}

\usepackage{color, soul}
\begin{document}

\maketitle
\thispagestyle{empty}
\pagestyle{empty}

\begin{abstract}
Manipulation of articulated and deformable objects can be difficult due to their compliant and under-actuated nature. Unexpected disturbances can cause the object to deviate from a predicted state, making it necessary to use Model-Predictive Control (MPC) methods to plan motion. However, these methods need a short planning horizon to be practical. Thus, MPC is ill-suited for long-horizon manipulation tasks due to local minima. In this paper, we present a diffusion-based method that guides an MPC method to accomplish long-horizon manipulation tasks by dynamically specifying sequences of subgoals for the MPC to follow. Our method, called Subgoal Diffuser, generates subgoals in a coarse-to-fine manner, producing sparse subgoals when the task is easily accomplished by MPC and more dense subgoals when the MPC method needs more guidance. The density of subgoals is determined dynamically based on a learned estimate of reachability, and subgoals are distributed to focus on challenging parts of the task. We evaluate our method on two robot manipulation tasks and find it improves the planning performance of an MPC method, and also outperforms prior diffusion-based methods.

More visualizations and results can be found at \href{https://sites.google.com/view/subgoal-diffuser-mpc}{https://sites.google.com/view/subgoal-diffuser-mpc}

\end{abstract}

\section{Introduction}

Robotic manipulation of articulated and deformable objects is challenging in part because they are compliant and under-actuated. External forces from the environment, e.g. friction from contact, or other disturbances can cause the actual state of the object to deviate from that predicted by a simulator. Thus, it is necessary to adapt to disturbances quickly when manipulating these objects. Sample-based Model-predictive Control (MPC) methods~\cite{williams2016aggressive, rubinstein2004cross} are a good choice for this application due to their flexibility, as they do not impose stringent requirements on the form of the cost function and dynamics.

However, these methods trade off horizon length in favor of speed, making them ill-suited for long horizon manipulation tasks. This paper presents an approach to robotic manipulation of articulated and deformable objects that overcomes this limitation by using a learned conditional generative model to dynamically predict sequences of subgoals. These subgoals are then used as guidance by the MPC method. 

Recent work on learning conditional generative models for manipulation has produced powerful methods based on diffusion~\cite{ho2020denoising}. These methods demonstrate the capacity to capture the distribution of states and actions required to produce trajectories for certain manipulation tasks~\cite{janner2022planning, ajay2022conditional, chi2023diffusion}. However, they either output the full trajectory directly~\cite{janner2022planning, ajay2022conditional, chi2023diffusion}, or adopt a fixed hierarchical structure~\cite{li2023hierarchical}. Also, they use a learned policy for low-level control, which can be data-inefficient and does not generalize well to new situations.

In this paper we present a method to generate subgoals using a diffusion model and delegate the task of finding a sequence of controls to move between subgoals to an MPC method. For these subgoals to be effective, we require the ability to produce subgoals at different resolutions (in terms of the number of steps needed to move between them). This is crucial for accomplishing difficult manipulation tasks, as varying levels of guidance are needed at different times. For example, consider the task of picking up an open notebook from the floor and placing it, closed, on a table~(Fig.~\ref{fig:subgoal_vis}). Transporting the notebook to the table may require sparse subgoals and minimal guidance for MPC, whereas placing the notebook down and folding it requires more careful manipulation.
In order to generate a sequence of subgoals at an appropriate resolution, we propose a diffusion-based architecture called Subgoal Diffuser. This method
generates subgoals in a coarse-to-fine manner. It initially outlines a coarse high-level plan with subgoals spaced far apart and subsequently fills in more subgoals as necessary.

 We introduce a reachability-based measure to determine when it is necessary to add more subgoals.
The main idea is that more subgoals should be used if adjacent subgoals are not reachable given the low-level MPC controller. Reachability is learned from the same dataset that is used to train the diffusion model. 
Furthermore, our method uses this learned distance metric to dynamically redistribute the subgoals to focus on the challenging parts of the task. Thus, the contributions of this paper are:
\begin{itemize}
    \item A diffusion-based framework to generate subgoals in a coarse-to-fine manner.
    \item A strategy based on an estimate of reachability to determine a suitable subgoal resolution for the task.
    \item A system that integrates subgoal generation and MPC for robot manipulation.
\end{itemize}

Our experiments on notebook and rope manipulation show that the generated subgoals effectively prevent the myopic MPC controller from falling into local minima. Our method also compares favorably to existing diffusion-based methods.

\section{Related Work}

\subsection{Diffusion models for Robotics}

Diffusion models have shown great premise in generative modeling, such as images~\cite{saharia2022photorealistic} and videos~\cite{ho2022imagen}. Recently, researchers have applied diffusion models to various robotics applications, such as data augmentation~\cite{yu2023scaling, bharadhwaj2023roboagent}, grasp synthesis~\cite{urain2023se}, text-conditioned scene rearrangement~\cite{kapelyukh2023dall, liu2022structdiffusion}, constrained trajectory generation~\cite{power2023sampling}, and motion planning~\cite{carvalho2023motion}. In this paper, we focus on robot manipulation. Diffuser~\cite{janner2022planning} and SceneDiffuser~\cite{huang2023diffusion} propose to jointly model the dense trajectory of states and actions, and draw the connection with standard trajectory optimization techniques. Another line of work~\cite{pearce2023imitating, chi2023diffusion, reuss2023goal} explores using diffusion models in the context of imitation learning and only models the distribution of demonstrated actions. 
Ajay et al.~\cite{ajay2022conditional} and Li et al.~\cite{li2023hierarchical} train a state-based diffusion model to predict desirable future states and a low-level policy to reach the predicted states. In contrast, we introduce a hierarchical framework for modeling the distribution of states (subgoals), and a procedure to automatically determine subgoal resolution required for the task. Also, we leverage MPC for low-level control. In our experiments, we show that dynamically deciding the subgoal resolution is critical for the task performance.

\subsection{Subgoal generation for long horizon planning} 
Prior work has used reachability to decompose long-horizon tasks. Hierarchical Visual Forsights (HVF)~\cite{nair2019hierarchical} proposes to estimate the reachability between adjacent subgoals by explicitly running an MPC method to plan. However, it requires running MPC on multiple start-goal pairs, which is computationally expensive. Other methods~\cite{eysenbach2019search,nasiriany2019planning,fang2022planning} leverage learning to estimate reachability. They first train a goal-conditioned policy using reinforcement learning, then they frame subgoal generation as online optimization over the value function of the policy. While effective, like other RL methods, they are sample-inefficient and require online interaction.
DiffSkill~\cite{lin2022diffskill} follows a similar strategy but avoids the caveats of RL by training the policy with demonstrations obtained by trajectory optimization. We propose a way to evaluate the reachability of an MPC controller that does not require demonstrations or online interaction. 

\subsection{MPC with a learned prior}
Learning a prior distribution of actions and subgoals has been used to speed up MPC and accomplish  complex tasks. Power and Berenson~\cite{power2022variational} leverage normalizing flow for modeling the action distributions. Wang and Ba~\cite{wang2019exploring} use a policy network to initialize the action sequences for MPC.
Sacks and Boots~\cite{sacks2022learning} introduce a framework with a learned optimizer with imitation learning, which makes better use of the expert samples. Similar to us, Li et al.~\cite{li2021mpc} propose an MPC framework with a generator for intermediate waypoints and a discriminator to choose the best waypoint candidate. However, they all require expert demonstration. While some prior works~\cite{carius2022constrained, lai2022parallelised, ichter2018learning} do not require expert demonstrations, they cannot conduct global reasoning.
Our proposed method is able to generate subgoals to guide MPC to alleviate local minima, while only using a low-quality offline dataset.

\section{Preliminaries}
\begin{figure*}[h]
    \centering
    \includegraphics[width=\linewidth]{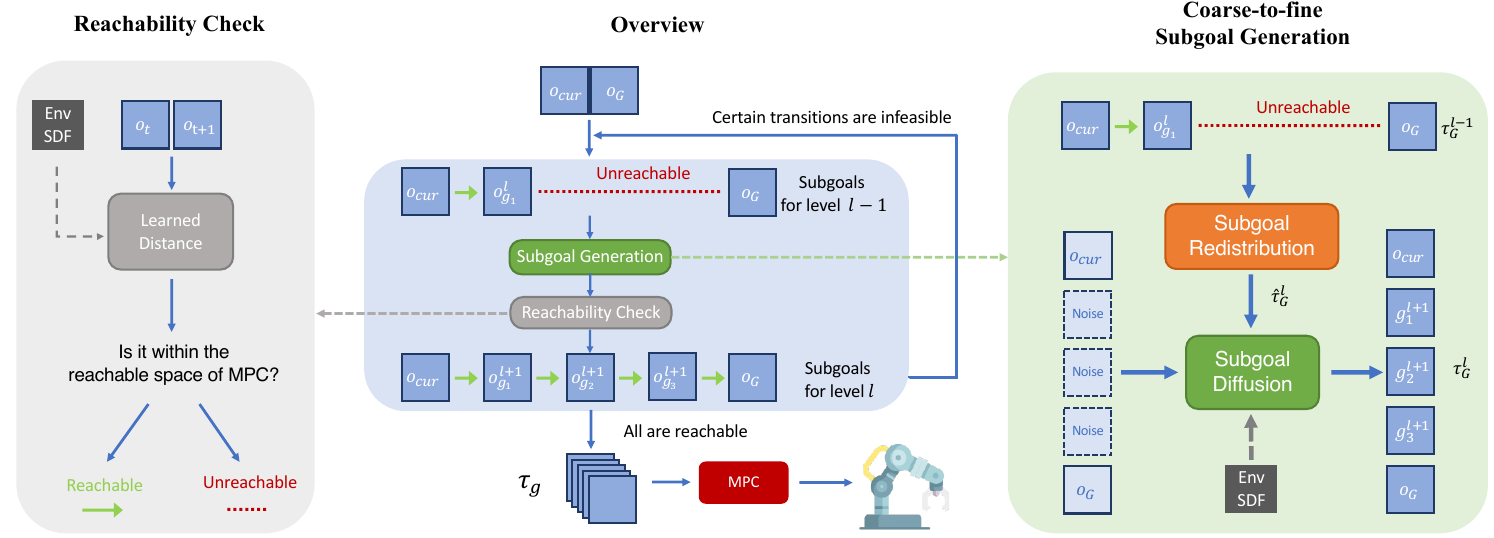}
    \caption{\textbf{Middle}: Our system consists of a diffusion model that generates subgoals in a coarse-to-fine manner, and a low-level MPC controller that tracks the subgoals. The diffusion model generates subgoals recursively until all subgoals are reachable from their predecessors. \textbf{Left}: To estimate the reachability, we learn a function that estimates the number of steps required to move between the two subgoals. If the prediction is smaller than the horizon of the MPC, we assume it is reachable. \textbf{Right}: Our Subgoal Diffuser is conditioned on current state, goal state, subgoals from the previous level, and (optionally) an SDF of the environment. The subgoals from the previous level will be redistributed so that they are equally spaced in terms of execution steps.}
    \vspace{-5mm}
    \label{fig:sys_fig}
\end{figure*}
\subsection{Problem Statement}
\label{sec:problem_statement}
We consider the problem of discrete-time optimal control with state denoted by \( \bx_t \in \mathbb{R}^{d_\bx} \) and control by \( \bu_t \in \mathbb{R}^{d_\bu} \). The state consists of two components: robot state $\br_t \in \bR$ and object state $\bo_t \in \bO$. After applying the control, the system will transition to the next state with a known transition probability function represented as \( \bx_{t+1} = f(\bx_t, \bu_t) \). A trajectory of states is defined as $\btau_{\bx} \triangleq [\bx_0 , \bx_1 , \dots , \bx_{T-1}]$, and a trajectory of controls is defined as $\btau_\bu \triangleq [\bu_0 , \bu_1 , \dots , \bu_{T-1}]$. Thus, the full trajectory is denoted by $\btau = [\btau_\bx ;\btau_\bu]. $ Given a cost function $\bJ$, MPC seeks to find a sequence of controls $\btau_\bu$ of length $H$ (the planning horizon) that minimizes $\bJ$. 

In this paper, we mostly consider the manipulation problem of under-actuated objects, such as rope.
Our goal is to find a sequence of robot controls $\btau_\bu$ to move the object from the current state $\bo_{cur}$ to a the desired configuration $\bo_{G}$. The cost function $\bJ$ is a function that measures distance to the goal state, e.g., Euclidean distance.

We assume the structure of the environment is provided in the form of Signed Distance Function (SDF), and the 3D model of the object is known.

This type of manipulation problem is challenging since the state space is high-dimensional and the object is under-actuated. 
We do not assume that high-quality demonstrations of the task are available.
To tackle this problem, we resort to a data-driven approach where we assume access to an offline dataset $\gD \triangleq \{\btau^i\}_{0\leq i < N} $ , which contains robot interactions with the target object. The dataset is collected by a random policy. 
Our goal is to learn a generative model that is able to produce a sequence of subgoals for the object: 
\vspace{-.3mm}
\begin{equation}
    \btau_{\mathcal{G}}=[\bo_{cur} , \bo_{g_1} , \dots , \bo_{g_{M-2}}, \bo_{G}]
\end{equation}
\noindent given current state, goal state, and (optionally) a scene representation. The subgoals will be used to guide a sampling-based MPC method to complete the task. The number of subgoal $M$ is variable and will be automatically estimated based by our method.

\subsection{Diffusion Models}
Diffusion models~\cite{sohl2015deep, ho2020denoising} are a powerful class of generative models designed to approximate the data distribution $q(\btau_0)$ from the dataset $ \gD \triangleq \{\btau^i\}_{0\leq i < M}$. Diffusion models frame the data generation as a $K$-step iterative denoising procedure, with a predefined forward noising process $q(\btau_{k+1}|\btau_{k})=\gN(\btau_{k+1};\sqrt{\alpha_k}\btau_k,(1-\alpha_k)\bI)$, and a learnable reverse denoising process $p_\theta(\btau_{k}|\btau_{k+1})$. The forward diffusion process can be seen as gradually fusing data with noise, and $K$ and $\alpha$ are hyperparameters that define this noise schedule. The data distribution $p_\theta(\btau_0)$ is expressed as:
\begin{equation}
    p_\theta(\btau_0) = \int p(\btau_K) \prod_{k=1}^K p_\theta(\btau_{k-1}|\btau_{k}) d \btau^{1:K}
\end{equation}
where $p(\btau_K)$ is a unit Gaussian prior. In practice, the data generation process is usually implemented via stochastic Langevin Dynamics~\cite{welling2011bayesian} starting from Gaussian noise.

While diffusion models can by trained by optimizing the variational lower-bound on $\log p_\theta(\btau)$, like prior work~\cite{janner2022planning,ajay2022conditional,chi2023diffusion}, we use the simplified objective from DPPM~\cite{ho2020denoising}:
\begin{equation}
\label{eq:diffusion_obj}
    \gL_{denoise}(\theta)=\mathbb{E}_{k\sim[1,K],\btau_0\sim q, \epsilon\sim \gN(\boldsymbol{0},\bI)}||\epsilon - \epsilon_\theta(\btau_k, k)||^2
\end{equation}
where $\epsilon_\theta$ is parameterized by a neural network to estimate the noise that can be used to recover the original data. 

\section{Method}

 We propose to generate a sequence of subgoals  $\btau_{\mathcal{G}}$ to guide a sampling-based MPC method to perform a manipulation task. In Sec.~\ref{sec:subgoal_gen}, we describe \emph{Subgoal Diffuser}, which generates a sequence of subgoals recursively from coarse to fine. With \emph{Subgoal Diffuser}, we can generate an arbitrary number of subgoals. 
 However, it is not clear how to determine how much guidance should come from the subgoals and how much should be left up to the MPC method. For example, tasks that are temporally extended or sensitive to error may require finer reasoning and thus more subgoals. To address this problem, we introduce a reachability-based method to dynamically determine the number of subgoals required for the task (Sec.~\ref{sec:adaptive_res}). Then we discuss how we integrate it with an MPC controller (Sec.~\ref{sec:method:mpc}) as well as the implementation details (Sec.~\ref{sec:method:implementation}).

\subsection{Coarse-to-fine Subgoal Generation using Diffusion}
\label{sec:subgoal_gen}
For challenging problems such as rope manipulation, generating a full sequence of locally and globally coherent subgoals in one shot is difficult. 
In this section, we introduce a diffusion architecture that is able to generate a chain of subgoals in a coarse-to-fine manner
to enable planning at different temporal resolutions. 

First, let $\Delta t$ be the \textit{temporal resolution}, which is the number of time steps between two consecutive states in a trajectory. An object trajectory $\tau_o = [\bo_0, \bo_1, \dots , \bo_{T-1}]$ of length $T$ has temporal resolution $\Delta t = 1$ between each pair of consecutive object states. When we set $\Delta t > 1$, we are able to extract a sequence of $\rmM$ subgoals $\btau_\mathcal{G}=[\bo_{g_0}, \bo_{g_1}, \dots,\bo_{g_{M-1}}]$, such that $\bo_{g_i}=\bo_{i * \Delta t}$ and $(\rmM-1) * \Delta t = T$. We define $\bo_{cur}=\bo_{g_0},\bo_G=\bo_{g_{M-1}}$ and the hierarchy starts at $M_0=2$. To enable coarse-to-fine subgoal generation, we define a hierarchy of $L+1$ levels of subgoals $\btau_\mathcal{G}^{0:L}$. Level $l$ contains $M_l$ subgoals, and $\rmM_{l+1} = 2\times \rmM_{l} -1$. 

To model $p(\btau_\mathcal{G}|\bo_{cur}, \bo_G)$ at different resolutions, we propose a novel architecture - \emph{Subgoal Diffuser}, which generates subgoals in a coarse-to-fine manner. As shown on the right side of Fig.~\ref{fig:sys_fig}, \emph{Subgoal Diffuser} is a conditional generative model $p_\theta(\btau_\mathcal{G}^l|\bo_{cur}, \bo_G,\btau_{\mathcal{G}}^{l-1})$ that predicts finer subgoals given current state, goal state and subgoals generated from the previous level. $\btau_\mathcal{G}^L$ can be predicted in a recursive manner:
\vspace{-1mm}
\begin{equation*}
    p(\btau_\mathcal{G}^{L}|\bo_{cur}, \bo_G) = p(\btau_\mathcal{G}^0|\bo_{cur}, \bo_G) \prod_{l=1}^{L} p_\theta(\btau_\mathcal{G}^{l}|\bo_{cur}, \bo_G,\btau_{\mathcal{G}}^{l-1})
\end{equation*}
\vspace{-1mm}

At each level, the subgoals are predicted in an in-painting manner~\cite{janner2022planning, sohl2015deep}. As shown on the right side of Fig.~\ref{fig:sys_fig}, the subgoal chain is initialized with Gaussian noise and gradually denoised into plausible subgoals during the reverse diffusion process. The first and final subgoals are $\bo_{cur}$ and $\bo_{G}$, which serve as conditioning. They are kept fixed throughout the diffusion process.

\begin{figure}
    \vspace{-0mm}
    \centering
    \includegraphics[width=0.9\linewidth]{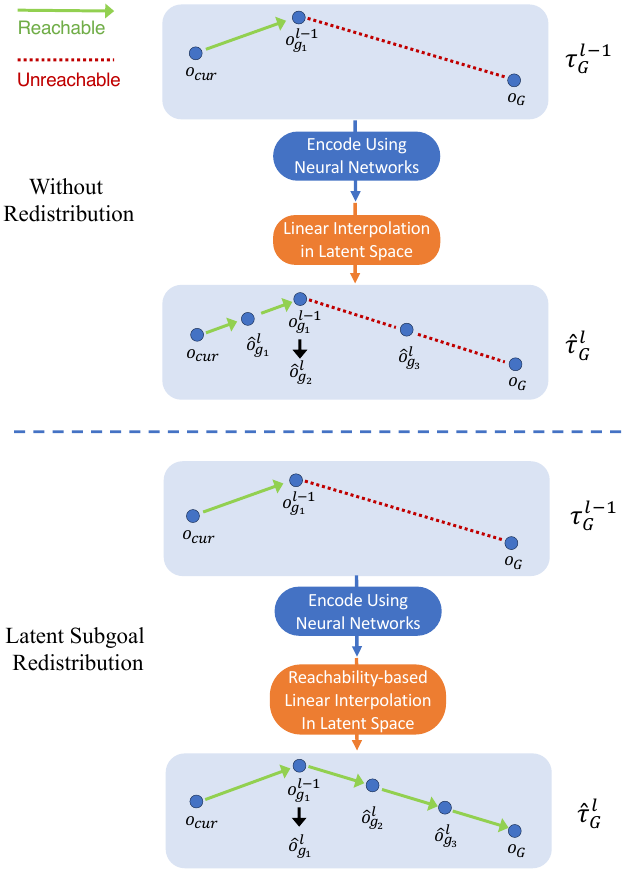}
    \caption{
    Prediction of finer subgoals $\btau_{\mathcal{G}}^{l}$ are generated contitioned on the coarse subgoals $\btau_{\mathcal{G}}^{l-1}$. To compute the conditioning $\hat{\btau}_{\mathcal{G}}^{l}$, we encode $\btau_{\mathcal{G}}^{l-1}$ into latent space and upsample it using linear interpolation. \textbf{Top}: Without redistribution, the new subgoals are evenly distributed, ignoring the relative distance between consecutive subgoals. Thus, the subgoals that are far apart remain unreachable. \textbf{Bottom}: $\btau_{\mathcal{G}}^{l-1}$ are redistributed in latent space using the estimated pairwise distance. By doing so, more subgoals will be filled in to the unreachable segment.}
    \vspace{-6mm}
    \label{fig:subgoal_redist}
\end{figure}

\emph{Subgoal Diffuser} is also conditioned on the predicted subgoals of the previous level $\btau_{\mathcal{G}}^{l-1}$. Since higher-level subgoals $\btau_{\mathcal{G}}^{l-1}$ are coarser than $\btau_{\mathcal{G}}^{l}$, we need a way to ``upsample'' them to a higher resolution. A straightforward way is to upsample subgoals via linear interpolation under the assumption that\emph{ the generated subgoals are equally spaced} (top figure in Fig.~\ref{fig:subgoal_redist}). However, this assumption does not always hold, especially for long-horizon problems. MPC may require more steps between certain pairs of subgoals than others, thus requiring more subgoals to be generated in-between. 

To account for this, we propose to upsample $\btau_{\mathcal{G}}^{l-1}$ according to the pairwise reachability between adjacent subgoals. Reachability is estimated by a learned function that approximates the number of steps the MPC will take to reach the next subgoal (described in Sec.~\ref{sec:adaptive_res}). By doing so, the model focuses more on connecting the distant subgoals, which reduces the chance that a myopic MPC method becomes stuck in a local minima (illustrated in bottom plot of Fig.~\ref{fig:subgoal_redist}). The assumption is that the MPC method is more likely to be stuck when subgoals are farther away. Also, since linear interpolation in the object state space $\bO$ could be problematic, e.g. creating unrealistic states, we encode $\btau_{\mathcal{G}}^{l-1}$ into a latent space using a neural network $f_\phi$, then interpolate $f_\phi(\btau_{\mathcal{G}}^{l-1})$ according to the reachability estimate and obtain a higher resolution subgoal chain $\hat{\btau}_{\mathcal{G}}^{l}$. 

We use a diffusion model with a temporal U-Net architecture, similar to~\cite{janner2022planning,carvalho2023motion}. Temporal U-net applies a 1D convolution over the time dimension and allows for generating different number of subgoals using a single model. 

Optionally, our method can also be conditioned on an SDF of the environment. We use two approaches for extracting information from the SDF: 1. \emph{Global conditioning}. We process the SDF using a 3D convolution neural network to condense the information of the entire scene into a single feature vector. 2. \emph{Local conditioning}. We render the point cloud of the object and compute the SDF value of each point. Then we process the point cloud with a PointNet ++~\cite{qi2017pointnet++} model to extract local contact information. 

\subsection{Adaptive Subgoal Resolution Selection}
\label{sec:adaptive_res}
\begin{wrapfigure}{r}{0.23\textwidth}
    \vspace{-0mm}
    \centering
    \includegraphics[width=\linewidth]{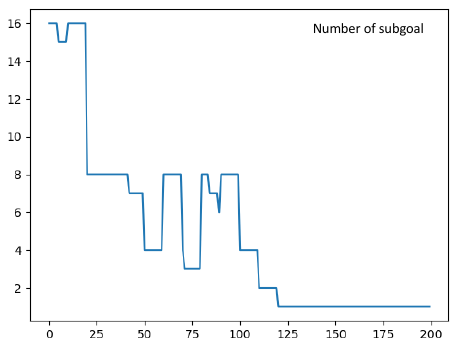}
    \caption{Number of subgoals as the robot progresses.}
    \vspace{-3mm}
    \label{fig:real_robot}
\end{wrapfigure}
Now we have a subgoal generator that is able to generate an arbitrary number of subgoals, but  the method still needs to decide how many subgoals are necessary to complete the task. 
When the number of subgoals is insufficient, the low-level MPC controller will fail to follow the subgoals and get trapped. When there are too many subgoals, the distances between subgoals become small, and the MPC controller will make limited progress when tracking each subgoal. In our experiments, (Sec.~\ref{sec:exp:sim}) we found that this is the reason why the baselines~\cite{ajay2022conditional, chi2023diffusion} got stuck in later stages of the task.

Our insight is that the number of subgoals is sufficient when the MPC controller can travel between subgoals successfully, i.e., all subgoals are reachable from their predecessors. Since we focus on an offline learning setting, we cannot learn a reachability function for an MPC controller through online trial-and-error. Instead, we assume that our MPC controller is able to reach $\bo_b$ from $\bo_a$, if the random policy that we use to collect the dataset can. Put another way, $\bo_b$ is reachable from $\bo_a$, if in the offline dataset there exists a path $\btau_{a\rightarrow b}$ that goes from $\bo_a$ to $\bo_b$ with steps $k<H$, where $H$ is the horizon of the MPC controller. 

To estimate the least number of steps $k$ required to travel between $
\bo_a$ and $\bo_b$ in the dataset, we follow \cite{hejna2023distance}. First, we model the whole distribution of travel time between states $p_{\psi}(k|\bo_a,\bo_b)$. Then, we select the smallest $k$ such that $p_{\psi}(k|\bo_a,\bo_b)>0$. To learn $p_{\psi}(k|\bo_a,\bo_b)$, we discretize the travel time into 40 bins, where the $i$-th bin represents $i$ steps, and the distances greater than 40 will be classified to be in the last bin. We frame this minimum distance estimation problem as classification, since regression will converge to the mean. To be robust to the error of function approximation, we use LogSumExp over the distribution to obtain a soft estimate of the least number of steps:
\begin{equation}
\label{eq:simple_lse}
\hat{d}(\bo_a,\bo_b) = -\alpha \log \E_{k \sim p_\psi(\cdot|\bo_a,\bo_b)}\left[e^{-k/\alpha}\right],
\end{equation}
where $\alpha$ is a temperature parameter that controls the softness of this estimate. We say that $\bo_b$ is reachable from $\bo_a$ if $\hat{d}(\bo_a,\bo_b) <H$. During test time, we use this learned metric to compute the length of all segments in the subgoal chain.
If the maximum $\hat{d}$ is larger than $H$, we increase the temporal resolution and generate more subgoals unless we have reached the maximum number of subgoals. $\hat{d}$ is also used during subgoal redistribution (Sec.~\ref{sec:subgoal_gen}).

\subsection{Model Predictive Control (MPC)}
\label{sec:method:mpc}
As shown in the middle figure of Fig.~\ref{fig:sys_fig}, \emph{Subgoal Diffuser} is integrated with an MPC controller to complete a robot manipulation task. We chose a sampling-based MPC method, MPPI~\cite{williams2016aggressive}, due to its robustness and flexibility. For our manipulation tasks, $\bu$ is the change in gripper position. Given a $\tau_u$ produced by MPPI, we roll it out in a simulator to produce the corresponding trajectory of object states $\tau_o$, which is used to evaluate cost. 

Usually, MPPI plans with a terminal cost for a single goal, yet simply picking the next subgoal predicted by diffusion does not perform well. This is because subgoal diffuser is trained on a random dataset, so the predicted subgoals can be sub-optimal (e.g. taking a detour), and it is safe to skip some intermediate subgoals. Therefore, we adopt the strategy of planning with goal sets, where MPPI considers all predicted subgoals simultaneously to compute the cost for a $\bo_t \in \tau_o$. The optimization problem that MPPI seeks to solve is then
\begin{align*}
    \argmin_{\tau_u} \sum_{t=0}^{H-1} &
    \Biggl( \min_{\bo_{g_i} \in \btau_{\mathcal{G}}} \left[||\bo_t-\bo_{g_i}||^2 - \lambda_{remote} \frac{1-\gamma^i}{1-\gamma}\right]\\ 
        &+\lambda_{col}\max(-SDF(\br_t),0)\\ 
        &+ \lambda_{smooth}||\bu_t-\bu_{t-1}||^2 \Biggr).
\end{align*}
The first term computes the distance to all subgoals with an incentive to encourage later subgoals in the chain, controlled by $\gamma$. The second term penalizes robot collisions (the simulator resolves object collisions since the objects are compliant), and the third term encourages smoothness of controls. 
A subgoal will be removed from the goal chain once it is reached.
We regenerate the subgoals $\btau_{\mathcal{G}}$ every 10 steps. 
We cannot guarantee that MPC will not become stuck when following the subgoal chains we predict. However, our results show that our method outperforms a baseline MPC method and two learning-based methods, suggesting that the subgoals we produce are indeed effective at guiding MPC.

\subsection{Implementation}
\label{sec:method:implementation}
The training dataset $\gD \triangleq \{\btau^i\}_{0\leq i < N} $ is collected using a random policy and contains 10,000 trajectories with a length of 100. The random policy will first sample a random reachable location in the free space and plan a collision-free trajectory to it using MPPI.
We define a subgoal hierarchy by specifying the number of subgoals at each layer $[M_0,M_1,\dots,M_{L-1}]$. We use [2, 3, 5, 7, 9, 17] in our experiments. In each training iteration, we sample a truncation of the trajectory $\hat{\btau}$ as well as the number of subgoals $M_l$. Then we subsample $M_l$ equally spaced states $\btau_\mathcal{G}^{l}$ as ground-truth subgoals and also $\btau_\mathcal{G}^{l-1}$ as conditioning. When training the diffusion model $p_\theta(\btau_\mathcal{G}^{l}|\bo_{cur},\bo_{G},\btau_{\mathcal{G}}^{l-1})$, we add Gaussian noise to $\btau_{\mathcal{G}}^{l-1}$ to approximate the prediction errors at test time. 
The model is trained according to Eq.~\ref{eq:diffusion_obj}.

During planning, MPPI samples 80 trajectories with a horizon of 10. We use a noise scale of 0.001 for action sampling and a temperature of 0.02 when computing the weights of sampled trajectories. We set $\lambda_{remote}=0.02$ and $\gamma=0.6$ to encourage the planner to reach later subgoals in the chain when possible. We also set $\lambda_{col}=10$ and $\lambda_{smooth}=0.001$. We warm-start the planner by initializing the nominal trajectory with results from the previous timestep. We run 5 iterations of refinement in the first step, and 2 iterations in the later steps. We use Mujoco~\cite{todorov2012mujoco} as the dynamics model for simulated and real-world experiments.
\section{Experiments}

\begin{figure*}
    \vspace{-0mm}
    \centering
    \includegraphics[width=\linewidth]{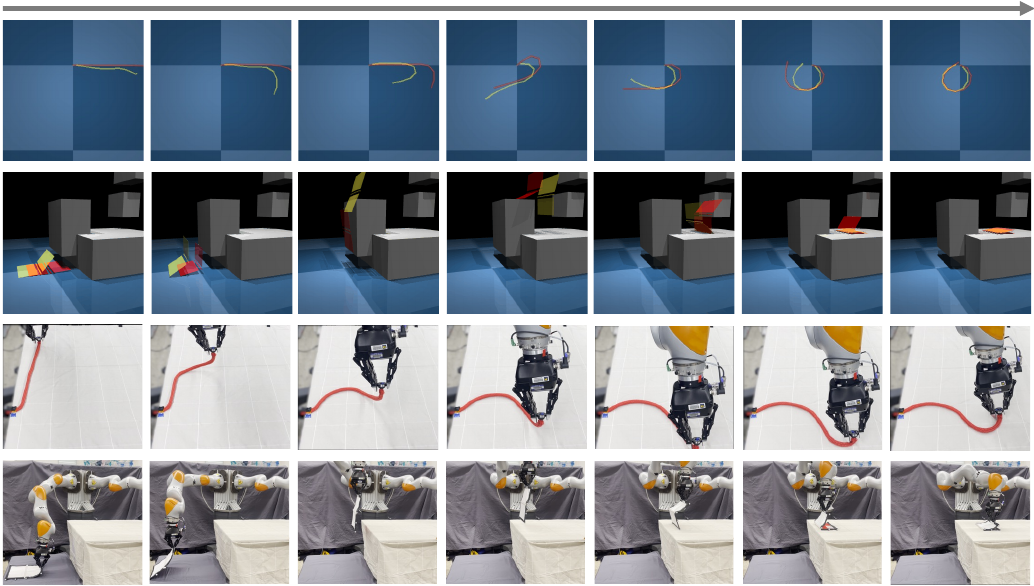}
    \vspace{-5mm}
    \caption{Snapshots of rope and notebook manipulation tasks in simulation (top two rows) and the real world (bottom two rows). The subgoals are visualized in yellow for simulation experiments. More visualizations can be found on our \href{https://sites.google.com/view/subgoal-diffuser-mpc/home}{website}}
    \vspace{-6mm}
    \label{fig:subgoal_vis}
\end{figure*}

\subsection{Simulated experiments}
\label{sec:exp:sim}
Our experiments seek to answer the following questions: (1) Can our method outperform the state-of-the-art diffusion-based methods? (2) What are the most important design choices in our method? We consider two difficult manipulation tasks in simulation.

\noindent\textbf{Rope reconfiguration}. Rope reconfiguration is a challenging manipulation task due to high-dimensional state space and complex dynamics. The goal of this task is to make certain shapes with rope, such as a circle or S shape. We attach one end of the rope to a fixed point and the other to a robot gripper. The rope is modeled as a 10-link linkage in Mujoco.

\noindent\textbf{Notebook manipulation}. To investigate how well the method generalizes to novel environments, we create an environment with randomized obstacles (as shown in Fig.~\ref{fig:subgoal_vis}). The goal is to pick up the notebook from the ground, lay it on the table, and close it while avoiding all the obstacles. This is a task that would intuitively be separated into several stages, yet it is unclear where the intermediate subgoals should be. The robot grasps the notebook in the middle of its edge.

For both tasks, we define the object state space $\bo$ using 10 keypoints on the object. All methods are evaluated on 10 start/goal pairs, and we use the euclidean distance to the goal as the evaluation metric. The maximum execution times are 350 (notebook) and 200 (rope) steps. 

\begin{table}
    \centering
    \begin{tabular}{@{}c|c|c@{}}
        \toprule
        Method & Rope $\downarrow$ & Notebook $\downarrow$  \\
        \midrule
        Ours &                                              $\mathbf{2.2\pm 0.9}$& $\mathbf{1.6 \pm 2.4}$\\
        Diffusion Policy~\cite{chi2023diffusion} &          $10\pm 6$ &   $7.9\pm 3.5$  \\
        Decision Diffuser~\cite{ajay2022conditional} &        $7.6\pm 1.7$  & $56\pm 9$\\
        MPPI &                                               $6.3\pm 5.4$ & $ 3.1 \pm 4.0$ \\
        \midrule
        Ours w/ fixed \# subgoals + receding horizon & $3.4\pm 1.8$& $3.1\pm 4.0$ \\
        Ours w/ fixed \# subgoals + fixed horizon & $3.6\pm 1.8$&  $8.7\pm12$\\
        Ours w/o coarse-to-fine & $2.5 \pm 0.7$ &  $9.5\pm 9.6$\\
        Ours w/o subgoal redistribution   & $2.6\pm 0.8$ & $2.4 \pm 3.4$\\
        \hline
    \end{tabular}
\caption{Mean and std. dev. of the  minimum distance to the goal over 10 test cases for each task.}
\label{table:sim_result}
\vspace{-6mm}
\end{table}

\vspace{-2mm}
\subsection{Baselines}
We compare our method to the following baselines and ablations:
\begin{itemize}
    \item Decision Diffuser~\cite{ajay2022conditional}: Decision Diffuser is the state-of-the-art offline reinforcement learning method. It models the dense trajectory of states by diffusion and extracts actions via an inverse dynamics model. During test time, it predicts a fixed-length trajectory.
    \item Diffusion Policy~\cite{chi2023diffusion}: Diffusion Policy is the state-of-the-art imitation learning method that directly models the action distribution of the dataset. Since our offline dataset contains low-quality data, we adapt the original implementation with hindsight relabeling~\cite{andrychowicz2017hindsight}, and add goal conditioning.
    
    \item Our method with fixed number of subgoals and receding horizon: In this baseline, we trained the model to always predict finest level of subgoals. During planning, a sequence of history states is used as conditioning, and the actual horizon is reduced as the history increases.
    \item Our method with fixed number of subgoals and fixed horizon: Similar to above, this variant also always predicts the finest level of subgoals. However, during planning, it only uses current state as conditioning so that the planning horizon is fixed.
    \item Our method without coarse-to-fine generation: In this baseline, the model is trained to predict $\btau_{\mathcal{G}}^{l}$ without being conditioned on $\btau_{\mathcal{G}}^{l-1}$. During planning, it also uses adaptive subgoal resolution selection.
   
    \item Our method without subgoal redistribution: For this baseline, we upsample $\btau_{\mathcal{G}}^{l}$  using linear interpolation and assume the subgoals are equally spaced.
\end{itemize}

For all methods, we record the minimum cost attained throughout the episode and compute their mean and standard deviation.
As shown in Table~\ref{table:sim_result}, our method outperforms all baselines for both tasks. Diffusion Policy~\cite{chi2023diffusion} doesn't work very well in our setting. This may be due to its' sensitivity to the quality of the dataset, even with Hindsight Relabeling. 

Decision Diffuser~\cite{ajay2022conditional} works slightly better but is still worse than our method. Since it predicts long, dense trajectories (H = 100), we find that it predicts very small actions when it is close to the goal and becomes stuck. 

MPPI alone is unable to solve complex manipulation tasks that contain local minima, thus obtaining sub-optimal performance. With the aid of our proposed subgoal generation method, the performance of MPPI is significantly improved.

We believe that part of the performance improvement over the Decision Diffuser and Diffusion Policy baselines comes from the fact that our method and MPPI use ground-truth dynamics models for planning. It is difficult to make a fair comparison, as diffusion policy and decision diffuser cannot be easily adapted to incorporate a ground-truth dynamics model. In fact, we consider the ability to leverage existing dynamics models to be an advantage of our method. 


Regarding ablations, compared against the two ablations with a fixed number of subgoals, we see a 30$\%$ performance gain by using adaptive resolution selection. The coarse-to-fine generation scheme and subgoal redistribution also help with the performance, especially on the Notebook task.

\vspace{-2mm}
\subsection{Physical Experiments}
To validate whether our method can be transferred to the real world, we replicated both the notebook manipulation and rope manipulation experiments using a 7 DoF Kuka LBR iiwa arm. 
To estimate the state of the object in the real world, we used motion capture for the notebook and CDCPD~\cite{wang2021tracking} for the rope. It is important to note that the dynamics model we use for physical experiments (Mujoco simulation) is only a rough approximation of real-world dynamics. Although the dynamics model we use is not accurate, i.e., we model the notebook as a rigid hinge while in the real world it is deformable, our method is able to reach the goal state reliably by regenerating the subgoals and replanning. See the accompanying video for executions.

\section{Conclusion}
We introduce the Subgoal Diffuser, a novel architecture that generates subgoals recursively to guide Model Predictive Control. We also propose a reachability-based subgoal resolution selection scheme to dynamically determine the number of subgoals based on the difficulty of the task. Our experiments show that these methods effectively guide MPC to perform difficult long-horizon manipulation tasks.




\bibliographystyle{IEEEtran}
\bibliography{main}

\end{document}